\documentclass[hess, manuscript]{copernicus}

\usepackage{booktabs}
\usepackage{soul}
\usepackage[utf8]{inputenc}
\usepackage[T1]{fontenc}
\usepackage{lineno}
\usepackage{tabularx}
\usepackage{longtable}

\nolinenumbers

\begin{document}

\title{Caravan MultiMet: Extending Caravan with Multiple Weather Nowcasts and Forecasts}

\Author[1]{Guy}{Shalev} 
\Author[1]{Frederik}{Kratzert}

\affil[1]{Google Research}

\correspondence{Guy Shalev (guysha@google.com)}
\date{\today} 

\runningtitle{Caravan MultiMet}
\runningauthor{Shalev et al.}

\firstpage{1}
\maketitle

\begin{abstract}
The Caravan large-sample hydrology dataset \citep{kratzert2023caravan} was created to standardize and harmonize streamflow data from various regional datasets, combined with globally available meteorological forcing and catchment attributes. This community-driven project also allows researchers to conveniently extend the dataset for additional basins, as done 6 times to date (see \url{https://github.com/kratzert/Caravan/discussions/10}).
We present a novel extension to Caravan, focusing on enriching the meteorological forcing data. Our extension adds three precipitation nowcast products (CPC, IMERG v07 Early, and CHIRPS) and three weather forecast products (ECMWF IFS HRES, GraphCast, and CHIRPS-GEFS) to the existing ERA5-Land reanalysis data. The inclusion of diverse data sources, particularly weather forecasts, enables more robust evaluation and benchmarking of hydrological models, especially for real-time forecasting scenarios. To the best of our knowledge, this extension makes Caravan the first large-sample hydrology dataset to incorporate weather forecast data, significantly enhancing its capabilities and fostering advancements in hydrological research, benchmarking, and real-time hydrologic forecasting. The data is publicly available under a CC-BY-4.0 license on Zenodo in two parts (\url{https://zenodo.org/records/14196771}, \url{https://zenodo.org/records/14196772}) and on Google Cloud Platform (GCP) -- see more under the Data Availability chapter.
\end{abstract}

\section{Introduction} \label{sec:introduction}

Large-sample hydrology (LSH) is crucial for understanding global water resources and addressing challenges like climate change impacts, flood and drought forecasting, and water management \citep{addor2020lsh}. Several LSH datasets have been instrumental in advancing hydrologic research. Notably, MOPEX \citep{schaake2006mopex} offered valuable hydrometeorological data for diverse catchments, contributing to model development and evaluation. \cite{newman2015camels} in combination with \cite{addor2017camels} with the release of the CAMELS dataset for the contiguous United States sparked a series of publications for regional LSH datasets. As of today, the existing dataset are CAMELS-AUS \citep{fowler2021camels}, CAMELS-BR \citep{chagas2020camels}, CAMELS-CH \citep{hoge2023camelsch}, CAMELS-CL \citep{alvarez2018camels}, CAMELS-DE \citep{loritz2023camelsde}, CAMELS-DK \citep{liu2024camelsdk}, CAMELS-FR \citep{camelsfr2024}, CAMELS-GB \citep{coxon2020camels}, and CAMELS-INDIA \citep{mangukiya2024camelsin}. Worth mentioning are also the LamaH-CE \citep{klingler2021lamah} and LamaH-ICE \citep{helgason2024lamahice} datasets, which in contrast to the CAMELS datasets additionally provide information about the gauge connectivity.  However, combining and comparing these datasets is not straightforward \citep{addor2020lsh}, as usually these regional datasets make use of high resolution, only locally available, forcing data and attribute maps. Here, the recent introduction of Caravan has marked a significant step forward -- by uniting and standardizing data from several large-sample hydrology datasets, and allowing community contributions to expand it to even more regions (see updating list here \url{https://github.com/kratzert/Caravan/discussions/10} including \cite{koch_2022_ext}, \cite{morin2023ext}, \cite{hoge2023ext}, \cite{casado2023ext}, \cite{helgason2024ext}, and \cite{farber2023ext}).

The forcing timeseries available in Caravan are derived from ERA5-Land \citep{munoz2021era5}. This was the product of choice due to its global coverage, relatively high quality, various surface variables, long records, availability on Earth Engine, and wide acceptance within the hydrologic community. While ERA5-Land has been invaluable for hydrologic modeling, it presents certain limitations -- as with any global product, the accuracy and reliability can vary spatially and temporally depending on the density and quality of the underlying observations. It has been shown that using a combination of several weather sources can increase performance of data driven hydrologic models (\cite{kratzert2023synergy}. This increases the motivation not to solely rely on a single product when possible, as also done operationally by Google's flood forecasting model (See \url{https://g.co/floodhub}).

Furthermore, relying solely on reanalysis data restricts hydrologic modeling to hindcasting scenarios. To advance hydrologic forecasting and to enable model intercomparison studies in the context of operational forecasting applications, it is essential to incorporate a historic archive of real-time weather forecast data. This allows researchers to evaluate model performance under realistic conditions, assess forecast uncertainty propagation through hydrologic models, and ultimately improve prediction accuracy for water resource management and hazard mitigation.

By including diverse weather data products and weather forecasts, this extension to Caravan bridges a critical gap in large-sample hydrology. It provides a unique platform for benchmarking hydrologic models in operational settings (\cite{nearing2024nature} and more recently \url{https://research.google/blog/a-flood-forecasting-ai-model-trained-and-evaluated-globally/}), enabling researchers to rigorously evaluate model performance across a range of meteorological forcings and forecast horizons. This capability is paramount for developing and deploying reliable hydrologic forecasting systems that can inform decision-making in real-time, ultimately enhancing our resilience to water-related challenges.

\section{Data}\label{sec:data}

Caravan includes meteorological forcing data from ERA5-Land, aggregated to daily resolution in the local time of each gauge location. You can find all details of the processing of the meteorological data in \cite{kratzert2023caravan}. In agreement with the existing data in Caravan, we spatially average all weather data across the catchment area and aggregate to daily resolution. However, for this extension, we keep all data in its original time (which is UTC+0) for multiple reasons: First, not all data products are available at sub-daily resolution, which is required to shift the data to local time, before computing daily statistics. Second, UTC+0 is the standard for global weather products, and by keeping the data of this extension in UTC+0, we make it easier to compare to additional products that are not included in this extension. For these reasons, our extension also includes daily aggregates of ERA5-Land, this time computed in UTC+0 and not in local time, as in the original Caravan dataset. 

Since not all of this data is available on Earth Engine, we process this data using internal Google infrastructure for all existing Caravan gauges, including all extensions. This approach has the limitation of not covering future community contributions. We aspire to mitigate this downside by periodical updates to the dataset.

\subsection{Structure of the dataset}\label{sec:data-structure}
For each nowcast or reanalysis product, we provide a single zarr file indexed by (1) basin and (2) date. And, for each forecast product, we provide a single zarr file which is indexed by (1) basin, (2) date, and (3) lead-time. The zarr format allows to efficiently read only parts of the dataset (e.g., only a subset of basins, dates, and lead-times) without the need to read all data into memory.

The data for the nowcast product is left-labeled, meaning that the value that appears under “date D”, refers to the time period [D 00:00:00, D+1 00:00:00].
For the forecast dataset, the date is the instant where the forecast was supposedly issued -- the fact these are sometimes reforecasts, and that operational forecasts have a few hours delay is ignored. The lead-times are expressed by datetime.timedelta(days=X). The first value is datetime.timedelta(days=1), and refers to the data in the range [D 00:00:00, D+1 00:00:00]; the second lead-time is indexed as datetime.timedelta(days=2), and refers to [D+1 00:00:00, D+2 00:00:00]; and so on.
For instance, if one would want to compare the data from all products for some date, they could query for date D, and additionally query for the first lead-time if it is a forecast product.

\subsection{Missing Data}\label{sec:missing-data}
The availability of each data-source was limited by the original data-sources, or the version of the data that Google possessed at the time. This can include missing dates (as in CHIRPS-GEFS), partial spatial cover (as in CHIRPS and CHIRPS-GEFS), and partial variables and years (as in HRES). We highlight the specific limitations in the relevant chapters below. Missing data-points are always denoted with nan values. In case a basin had no valid values for some weather product, it is removed (but will still exist for other products).

\subsection{Spatial Coverage}\label{sec:spatial-coverage}
\begin{figure}[ht]
    \centering
    \includegraphics[width=0.9\textwidth]{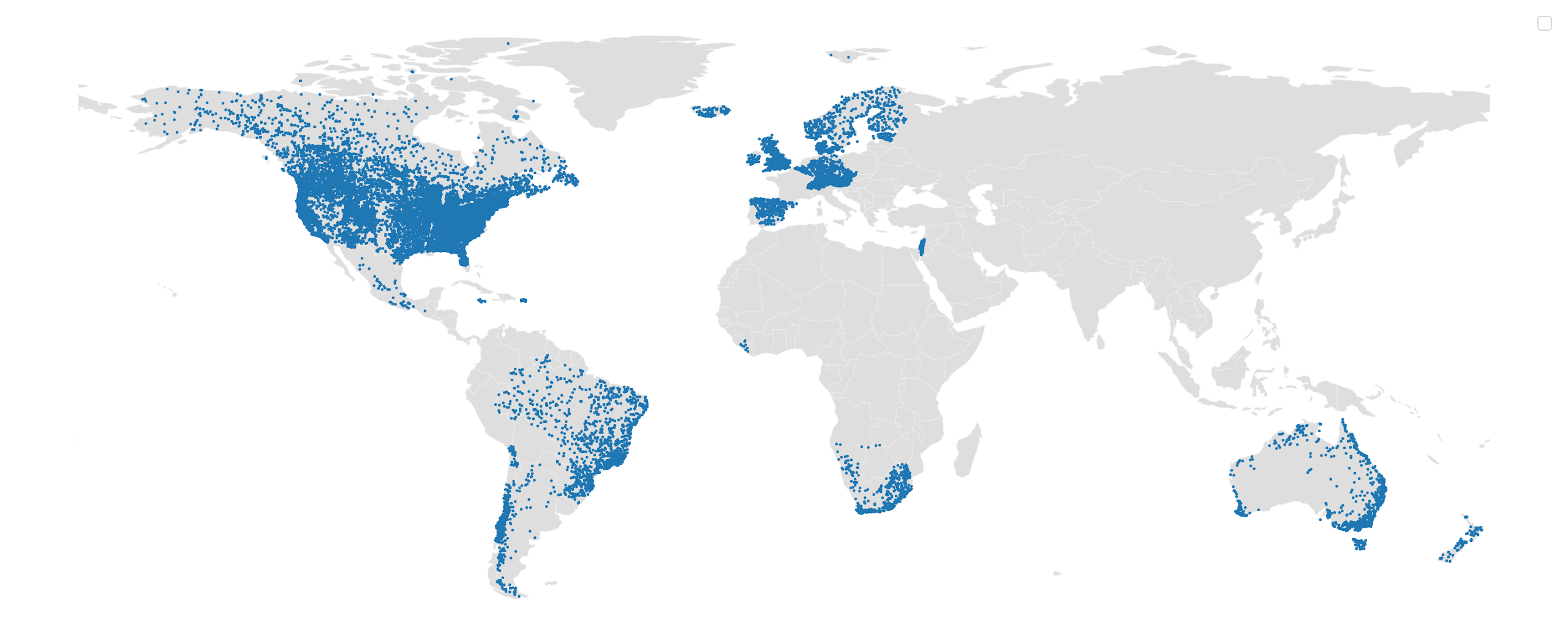}
    \caption{The locations of Caravan gauges covered by the Caravan MultiMet extension.}
    \label{fig:coverage-map}
\end{figure}

This extension covers 22,492 gauges included in the original Caravan dataset and its extensions, across 48 countries. See figure \ref{fig:coverage-map}.

\begin{table}[ht]
\centering
\begin{tabularx}{\textwidth}{|l|X|l|l|l|}
\hline
        Product & Variable & Units & Availability & Lead-times \\
\hline
CPC        & precipitation          & $mm$               & [1979-01-01, 2024-07-31]       & - \\
                 &   &    & with sporadic missing data              &  \\
\hline

IMERG v07 Early        & precipitation          & $mm$               & [2000-05-01, 2024-10-31]       & - \\
\hline

CHIRPS        & precipitation          & $mm$               & [1981-01-01, 2024-07-30]       & - \\
                 &   &    & within latitudes [-50°, 50°]              &  \\
\hline
ERA5-Land        &  dewpoint\_temperature\_2m          & $^{\circ}C$  & [1950-01-01, 2024-10-31]       &  - \\
                 &  potential\_evaporation\_DEPRECATED          & $mm$  &  &  \\
                 &  potential\_evaporation\_FAO\_PENMAN\_MONTEITH          & $mm$  &  &  \\
                 &  snow\_depth\_water\_equivalent          & $mm$  &  &  \\
                 &  surface\_net\_solar\_radiation          & $Wm^{-2}$  &  &  \\
                 &  surface\_net\_thermal\_radiation        & $Wm^{-2}$  &  &  \\
                 &  surface\_pressure          & $kPa$  &  &  \\
                 &  temperature\_2m          & $^{\circ}C$  &  &  \\
                 &  total\_precipitation          & $mm$  &  &  \\
                 &  u\_component\_of\_wind\_10m          & $ms^{-1}$  &  &  \\
                 &  v\_component\_of\_wind\_10m          & $ms^{-1}$  &  &  \\
                 &  volumetric\_soil\_water\_layer\_1          & $m^3/m^3$  &  &  \\
                 &  volumetric\_soil\_water\_layer\_2          & $m^3/m^3$  &  &  \\
                 &  volumetric\_soil\_water\_layer\_3          & $m^3/m^3$  &  &  \\
                 &  volumetric\_soil\_water\_layer\_4          & $m^3/m^3$  &  &  \\

\hline
CHIRPS-GEFS        & precipitation          & $mm$               & [2000-01-01, 2024-01-31]       & 1-16 days \\
                 &   &    & with some missing dates.              &  \\
                 &   &    & within latitudes [-50°, 50°]              &  \\
                 
\hline
ECMWF HRES       &  surface\_net\_solar\_radiation          & $Wm^{-2}$  & [2012-07-01, 2024-06-30]       & 1-10 days \\
                 &  surface\_net\_thermal\_radiation        & $Wm^{-2}$  &   &  \\
                 &  surface\_pressure          & $kPa$  &  &  \\
                 &  temperature\_2m          & $^{\circ}C$  &  &  \\
                 &  total\_precipitation          & $mm$  &  &  \\
\hline
GraphCast        &  temperature\_2m          & $^{\circ}C$  & [2016-01-02, 2023-12-21]       & 1-10 days \\
                 &  total\_precipitation          & $mm$  & 2022--2023 out-of-training &  \\
                 &  u\_component\_of\_wind\_10m          & $ms^{-1}$  & 0.25$^{\circ}$ longitude shift &  \\
                 &  v\_component\_of\_wind\_10m          & $ms^{-1}$  & (see Sec.~\ref{sec:graphcast}) &  \\
\hline

\end{tabularx}

\caption{Summary of all available data in Caravan MultiMet\label{tab:caravan-multimet-table}}
\end{table}

\section{Nowcast Products}\label{sec:nowcast-products}

\subsection{CPC}\label{sec:cpc}

The Climate Prediction Center (CPC) Unified Gauge-Based Analysis of Daily Precipitation is a gridded precipitation product generated by NOAA's Climate Prediction Center. This dataset is part of the CPC Unified Precipitation Project, which aims to produce a suite of high-quality, consistent precipitation products by integrating various data sources using optimal interpolation (OI) techniques \citep{cpc2007}.
The original temporal resolution is daily, and the spatial resolution is 0.5°. You can read more about the underlying dataset at \url{https://psl.noaa.gov/data/gridded/data.cpc.globalprecip.html} .

\begin{figure}[ht]
    \centering
    \includegraphics[width=0.6\textwidth]{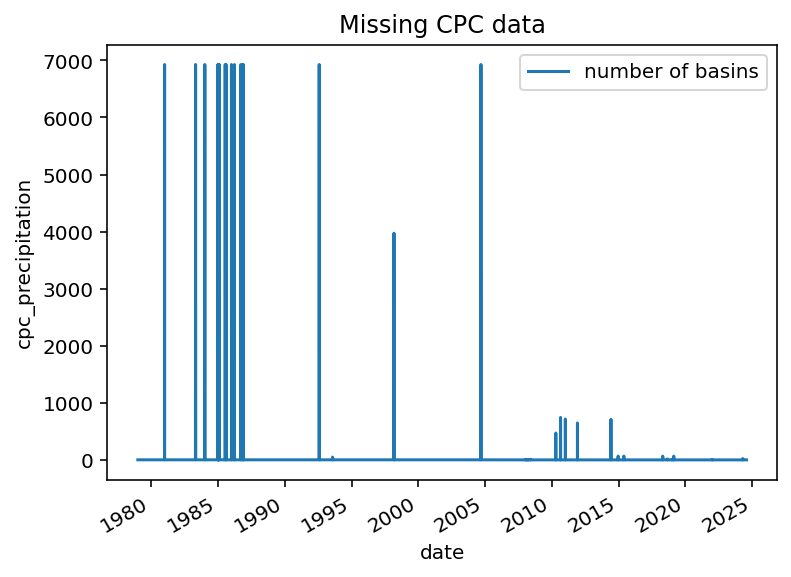}
    \caption{The number of basins with missing CPC data by date.}
    \label{fig:cpc-missing}
\end{figure}

The data provided in the Caravan-MultiMet extension covers the range of [1979-01-01, 2024-07-31]. Sporadic dates and locations where the data is missing in the original dataset are filled with nan values (see figure \ref{fig:cpc-missing}).

\subsection{IMERG}\label{sec:imerg}

The Integrated Multi-satellitE Retrievals for GPM (IMERG) is a NASA product estimating global surface precipitation rates at a spatial resolution of 0.1° every half-hour, starting in 2000. IMERG has three Runs with varying latency in response to a range of application needs: rapid-response applications (Early Run, 4-h latency), same/next-day applications (Late Run, 14-h latency), and post-real-time research (Final Run, 3.5-month latency). In this dataset we use the “Early” version of IMERG which is most applicable for time-sensitive real-time forecasting. Version 07 of IMERG was released in June 2024, and is to date the current version of the dataset. More information is available at:  

\url{https://data.nasa.gov/dataset/GPM-IMERG-Early-Precipitation-L3-Half-Hourly-0-1-d/mek8-8ssg/about_data}. The data was obtained from  \url{https://jsimpsonhttps.pps.eosdis.nasa.gov/imerg/early/}.

The data provided in the Caravan-MultiMet extension covers the range of [2000-05-01, 2024-10-31], with no missing data steps.

\subsection{CHIRPS}\label{sec:chirps}

CHIRPS (Climate Hazards group Infrared Precipitation with Stations) is a quasi-global high resolution, daily, precipitation dataset. Developed by the Climate Hazards Center at the University of California, Santa Barbara, CHIRPS blends satellite imagery with in-situ station data to create gridded rainfall estimates. More information is available at: \url{https://www.chc.ucsb.edu/data/chirps}.
The spatial resolution of the data is 0.05°, making this the highest spatial resolution data in this extension. Unfortunately, CHIRPS only covers the latitude range [-50, 50]. We provide the processed data for Caravan basins where the entire catchment is within this covered region.

The data provided in the Caravan-MultiMet extension covers the range of [1981-01-01, 2024-07-30], with no missing data steps in this range.

\subsection{ERA5-Land recomputed at UTC daily aggregation}\label{sec:era5}
ERA5-Land provides hourly high resolution information for several surface variables. You can read more about ERA5-Land and the way it is processed in the original Caravan paper.

We recompute the ERA5-Land data provided in the original Caravan dataset with one major difference -- the data is aggregated into daily resolution in UTC timezone for all basins, as opposed to the original dataset which computed the daily aggregation after shifting to the local timezone of each basin. This makes the ERA5-Land data in this extension compatible with the other meteorological datasets in Caravan-MultiMet, where it is not always possible to shift to the local timezone (e.g. in the case of native daily resolution products). It is important to note that the discharge data in Caravan is typically affected by the local timezone and does not necessarily match UTC. Another difference is that we only include daily averages (or daily sums, in the case of accumulating features of ERA5-Land), as opposed to the original Caravan dataset, which also provided the daily min and max values. 

Several studies raised concerns regarding the reliability of the ERA5-Land potential evaporation (\cite{clerc2024pet}, \cite{kratzert2023caravan}, \cite{klingler2021lamah}). We therefore add an additional potential evaporation variable, based on the FAO Penman-Monteith \citep{pet-guidelines} using other ERA5-Land variables, at a daily resolution, based on the code provided by \cite{singer2021fao} We name the new variable "potential\_evaporation\_FAO\_PENMAN\_MONTEITH", and include the original ERA5-Land variable (renamed to "potential\_evaporation\_DEPRECATED") to allow for their comparison.

The data provided in the Caravan-MultiMet extension covers the range of [1950-01-01, 2024-10-31], with no missing data steps for the provided basins.

\section{Forecast Products}\label{sec:forecasts}

\subsection{CHIRPS-GEFS}\label{sec:chirps-gefs}
\begin{figure}[ht]
    \centering
    \includegraphics[width=0.7\textwidth]{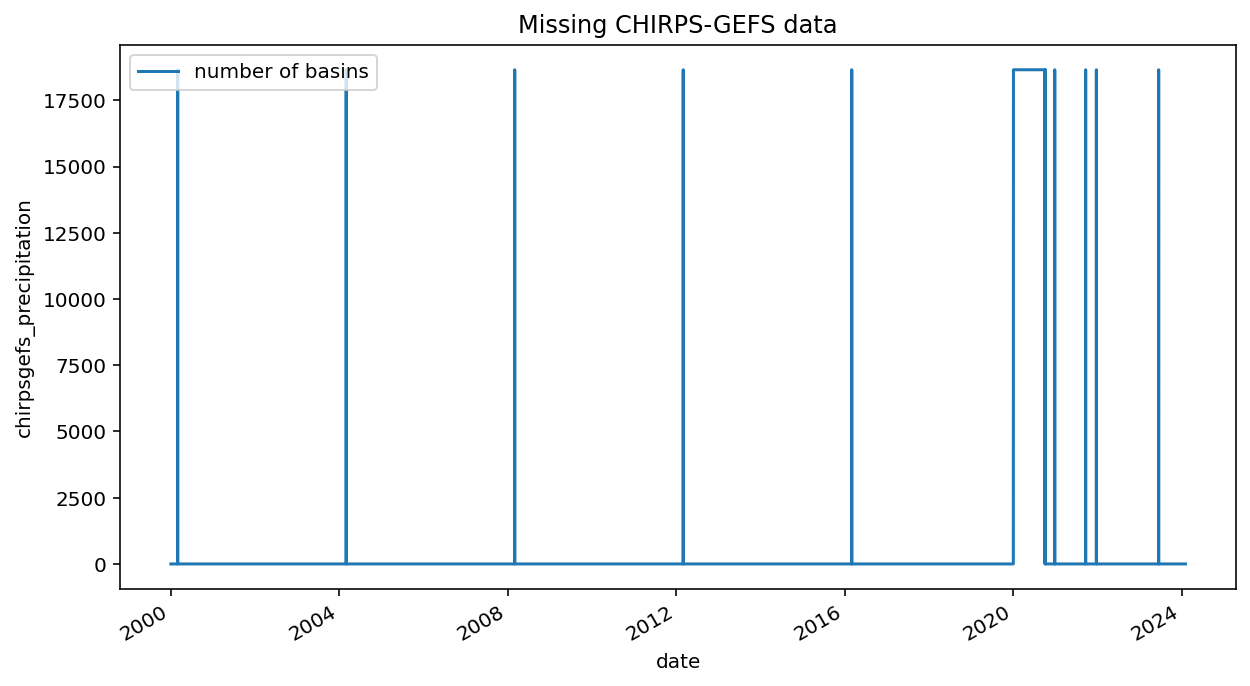}
    \caption{The number of basins with missing CHIRPS-GEFS data by date. Notably, a large part of 2020 is missing for all basins, and February 29th on leap years.}
    \label{fig:chirps-gefs-missing}
\end{figure}

CHIRPS-GEFS is a bias-corrected and downscaled version of NCEP Global Ensemble Forecast System (GEFS) version 12 precipitation forecasts made to be spatially compatible with Climate Hazards Center InfraRed Precipitation with Stations (CHIRPS) data.
It provides forecasts issued daily, with lead-times from 1 day to 16 days in daily resolution. The spatial resolution is 0.05° (as in CHIRPS). Unfortunately, as in CHIRPS, this product only covers the latitude range [-50, 50]. We provide the processed data for Caravan basins where the entire catchment is within this covered region.
The data provided in the Caravan-MultiMet extension covers the range of [2000-01-01, 2024-01-31], with some missing data steps (see figure \ref{fig:chirps-gefs-missing}).

\subsection{ECMWF IFS HRES}\label{sec:ecmwf-ifs-hres}

The European Centre for Medium-Range Weather Forecasts (ECMWF) Integrated Forecasting System (IFS) High-Resolution (HRES) model provides deterministic weather forecasts which include several variables. ECMWF’s global forecasts are recognised as amongst the most skilful in the world. It operates at an approximate spatial resolution of 0.1° and provides a weather forecast for the next 10 days.

Acquiring ECMWF’s forecast data is not as straightforward as other products. On the one hand, this makes this part of the extension more valuable, saving the user a lot of work acquiring the data. On the other hand, the limitations we had ourselves in querying the data affected the data we can responsibly provide in this dataset -- thus limiting us to only 5 variables, and only several years of data even though the full archive includes more.

The underlying data used to create this dataset is based on forecasts with issue time 00:00 UTC. The lead-times available are hourly for the first 90 hours; 3-hourly for hours 90-144; and 6 hourly onwards until 240 hours after the issue time. All of these are aggregated to daily values.

The surface variables provided in this extension are a subset of the ERA5-Land set of variables, and are in many ways compatible and comparable with that data -- see the list of variables and their units in Table \ref{tab:caravan-multimet-table}. The data provided in the Caravan-MultiMet extension covers the range of [2016-01-01, 2024-09-30], with no missing data steps.

\subsection{GraphCast}\label{sec:graphcast}
GraphCast is a machine learning (ML) model for global weather forecasting developed by Google DeepMind \citep{graphcast2023}. The model leverages graph neural networks (GNNs) to process spatially structured weather data and generate forecasts at a spatial resolution of 0.25$^{\circ}$, with 10 days lead time, at a 6-hourly temporal resolution.

Although the data is available for 4 issue-times per day, we use only the one produced at 00:00 UTC. The lead-times of GraphCast have a 6-hour density -- we aggregate the 4 lead-times of each day (by mean or sum, depending on the variable) to produce the daily lead-time density for 10 days.
The surface variables provided in this extension are precipitation, temperature, and wind (u10 and v10 components). Similarly to HRES, these too are compatible with the ERA5-Land data.

The data provided in the Caravan-MultiMet extension covers the range of [2016-01-02, 2023-12-21], with no missing data steps.

\subsubsection{Known issues and caveats (updated November 2026)}
After the initial release of Caravan MultiMet, two important issues were identified in the GraphCast subset (all other products in Caravan MultiMet are unaffected):

\begin{enumerate}
    \item \textbf{Training/test period split (only 2022 and 2023 are out-of-training test years):} An earlier version of this manuscript incorrectly stated that each year in 2016--2023 was generated using a GraphCast model trained up to that year. In fact, only \emph{two} HRES-fine-tuned GraphCast checkpoints were used to generate the archive, both pre-trained on ERA5 (1979--2017) and fine-tuned on ECMWF HRES with ERA5 6-hourly precipitation as target:
    \begin{itemize}
        \item \textbf{2016--2022:} Generated by a single checkpoint fine-tuned on HRES from 2016 through 2021.
        \item \textbf{2023:} Generated by a checkpoint that was also fine-tuned on 2022 (i.e., HRES from 2016 through 2022).
    \end{itemize}
    Consequently, \textbf{2016--2021 are in-training years}: 2016--2017 were included in both ERA5 pre-training and HRES fine-tuning, and 2018--2021 were included in HRES fine-tuning (including ERA5 precipitation targets). While these years can serve as historical forecast inputs when training downstream hydrological models, they must \emph{not} be used for out-of-training forecast evaluation. \textbf{Only 2022 and 2023 are out-of-training test years.}

    \item \textbf{Off-by-one longitude grid shift (0.25$^{\circ}$ westward):} Due to a data processing bug, the global $0.25^{\circ}$ GraphCast grid was shifted by one grid cell ($0.25^{\circ}$) westward prior to spatial averaging over catchment polygons. Consequently, each catchment received GraphCast forecast values from $0.25^{\circ}$ east of its true location.
\end{enumerate}

\subsubsection{Recommendation to use newer models}
Given the issues above -- and because deterministic GraphCast (also released as WeatherNext Graph, which was officially deprecated on July 29, 2026 and is no longer producing real-time forecasts) has been superseded by higher-skill probabilistic AI weather models -- we do not recommend using the Caravan MultiMet GraphCast subset to evaluate current AI weather forecasting capabilities. Instead, we encourage researchers to rely on newer probabilistic models, such as WeatherNext 3 (documented at \url{https://developers.google.com/weathernext/guides/models}).

\section{Data Verification}\label{sec:data-verification}
\subsection{ERA5-Land agreement with the original Caravan}
\begin{figure}[ht]
    \centering
    \includegraphics[width=0.9\textwidth]{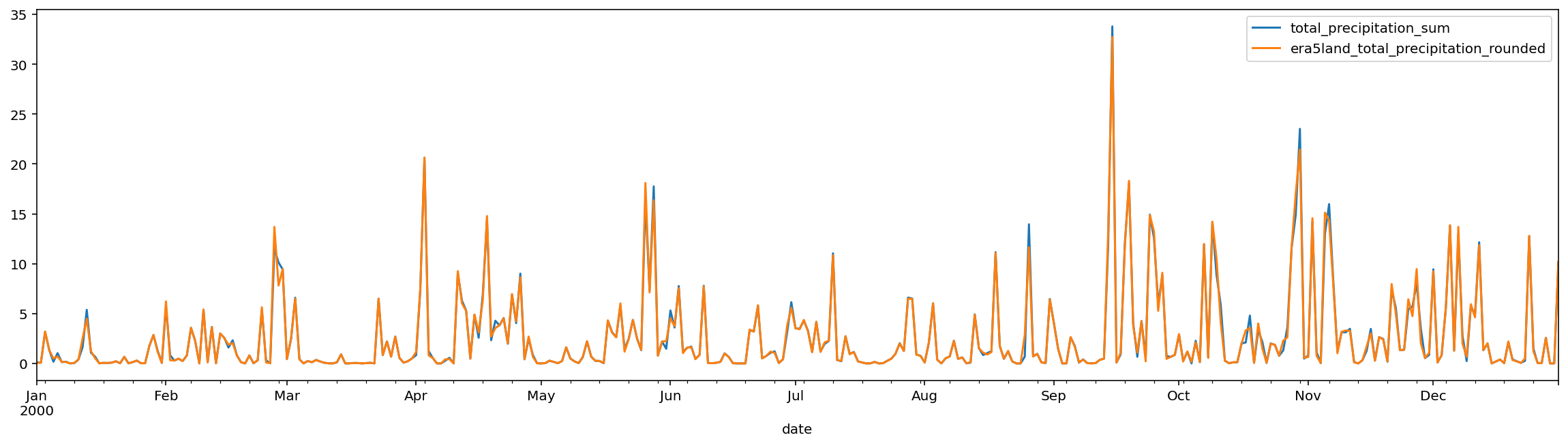}
    \caption{The ERA5-Land total precipitation for a year at the gauge "camelsgb\_54007", which shows strong agreement between the two versions of the data -- blue for the original Caravan dataset, orange for the MultiMet extension. It seems they have occasional disagreements on peak values.}
    \label{fig:precip-agreement}
\end{figure}

As a first step, we checked that our reprocessing of ERA5-Land agrees with the original dataset. Small and insignificant discrepancies are probably produced due to the processing code -- which is not entirely identical as the one used on Earth Engine. Since the original Caravan dataset was shifted to local timezone, we focus our analysis on an arbitrary gauge in Great Britain with UTC timezone, "camelsgb\_54007" in the year 2000. Note that the original Caravan also rounds the values to 2 decimal digits, while this extension doesn't. We apply the comparison a rounded version of the extension data. The results are shared in Table \ref{tab:r2-original-caravan}. Figure \ref{fig:precip-agreement} shows the data for total\_precipitation, the variable with the strongest disagreement ($R^2=0.988$) to get a feeling of the magnitude of such differences.

\begin{table}[ht]
\centering
\begin{tabular}{|l|c|} 
\hline
Variable & $R^2$ \\
\hline
dewpoint\_temperature\_2m\_mean  &  0.999  \\
\hline
temperature\_2m\_mean  &  0.999  \\
\hline
potential\_evaporation\_sum  &  0.999  \\
\hline
snow\_depth\_water\_equivalent\_mean  &  0.999  \\
\hline
surface\_pressure\_mean  &  0.998  \\
\hline
surface\_net\_solar\_radiation\_mean  &  1.0  \\
\hline
surface\_net\_thermal\_radiation\_mean  &  0.994  \\
\hline
volumetric\_soil\_water\_layer\_1\_mean  &  0.996  \\
\hline
volumetric\_soil\_water\_layer\_2\_mean  &  0.998  \\
\hline
volumetric\_soil\_water\_layer\_3\_mean  &  0.999  \\
\hline
volumetric\_soil\_water\_layer\_4\_mean  &  0.988  \\
\hline
total\_precipitation\_sum  &  0.989  \\
\hline
u\_component\_of\_wind\_10m\_mean  &  0.998  \\
\hline
v\_component\_of\_wind\_10m\_mean  &  0.998  \\
\hline

\end{tabular} 

\caption{$R^2$ score of all ERA5-Land variables for the year 2000 compared to the data in the original Caravan dataset. This is a verification that the data is in the correct units, and that the differences in the processing are of a reasonable size. This analysis is over the arbitrary gauge with UTC timezone "camelsgb\_54007"\label{tab:r2-original-caravan}}.
\end{table}

\subsection{Verification of other products}
Now that we have verified the ERA5-Land data of the extension agrees with the original dataset, we use it for reference and compare it to the other products. For each other variable in the other products, there is a trivially matching ERA5-Land variable (most commonly this is total\_precipitation, but for HRES and GraphCast several others exist). We compare each variable to their matching ERA5-Land variable, by applying $R^2$ score with ERA5-Land taken as the ground truth -- results are presented in Table \ref{tab:r2-shift-scores}. We also apply this test with temporal ($\pm 1$-day) shifts to the ERA5 data, to make sure we do not have a temporal off-by-one error in the process. Here, the maximal $R^2$ is obtained with no temporal shift for all variables. The reasonably high $R^2$ scores for the correct shift also provides verification that the units are correct, a property that is not captured by metrics such as correlation. The score was computed on all dates where both variables were available, and we include the number of samples used for the computation. Note that this check verifies temporal alignment only and did not detect the $0.25^{\circ}$ spatial longitude shift in GraphCast; additionally, GraphCast's $R^2$ scores across all 2,911 days include its 2016--2021 training years (see Section~\ref{sec:graphcast}). The results are presented in Table \ref{tab:r2-shift-scores}.

\begin{table}[ht]
\centering
\begin{tabularx}{\textwidth}{|l|l|l|l|X|}
\hline
        Variable & $R^2$ shift -1 & $R^2$ no shift & $R^2$ shift +1 & Num. valid samples for computation  \\
\hline
cpc\_precipitation & -0.29 & \textbf{0.49} & -0.08 & 16623 \\
\hline
imerg\_precipitation & -0.84 & \textbf{0.1} & -1.17 & 8919 \\
\hline
chirps\_precipitation & -0.39 & \textbf{0.14} & -0.69 & 15917 \\
\hline
chirpsgefs\_precipitation & -0.74 & \textbf{0.5} & -0.69 & 8512 \\
\hline
hres\_surface\_net\_solar\_radiation & 0.8 & \textbf{0.98} & 0.78 & 3196 \\
\hline
hres\_surface\_net\_thermal\_radiation & 0.66 & \textbf{0.97} & 0.64 & 3196 \\
\hline
hres\_surface\_pressure & 0.53 & \textbf{0.85} & 0.53 & 3196 \\
\hline
hres\_temperature\_2m & 0.96 & \textbf{1.0} & 0.96 & 3196 \\
\hline
hres\_total\_precipitation & -0.31 & \textbf{0.84} & -0.33 & 3196 \\
\hline
graphcast\_temperature\_2m & 0.95 & \textbf{0.98} & 0.94 & 2911 \\
\hline
graphcast\_total\_precipitation & -0.13 & \textbf{0.87} & -0.1 & 2911 \\
\hline
graphcast\_u\_component\_of\_wind\_10m & 0.24 & \textbf{0.87} & 0.16 & 2911 \\
\hline
graphcast\_v\_component\_of\_wind\_10m & 0.23 & \textbf{0.88} & -0.02 & 2911 \\
\hline

\end{tabularx}

\caption{$R^2$ score of all non-ERA5-Land variables with their matching ERA5-Land variable, with and without temporal ($\pm 1$-day) shifts. This is a verification that the data is not shifted in time, and that the different variables agree to a reasonable extent. Note that GraphCast values reflect all available days (including its 2016--2021 training period) and the $0.25^{\circ}$ spatial longitude shift described in Section~\ref{sec:graphcast}. This analysis was done for the sample gauge "GRDC\_6217400"\label{tab:r2-shift-scores}}
\end{table}

\section{Data and Code Availability}\label{sec:data-availability}

The data is available for download on Zenodo in two parts (\url{https://zenodo.org/records/14196771}, \url{https://zenodo.org/records/14196772}).

Additionally, a colab with a usage example of the dataset is available at \url{https://github.com/kratzert/Caravan/}. The colab reads the data from zarr files stored on GCP, which allows accessing only subsets of the data (specific variables, basins, dates, and lead-times) without the need to download the entire dataset and to read all of it into memory.

The Caravan-MultiMet dataset is released under CC-BY-4.0 license, like the majority of Caravan. The original license and source of each individual product are attached in the dataset itself. The original Caravan dataset is available at \url{https://doi.org/10.5281/zenodo.7540792} .

Any feedback, especially concerning issues with the data itself, will be highly appreciated.

\bibliographystyle{copernicus}
\bibliography{bibliography.bib}

\begin{thebibliography}{30}
\providecommand{\natexlab}[1]{#1}
\providecommand{\url}[1]{{\tt #1}}
\providecommand{\urlprefix}{URL }
\expandafter\ifx\csname urlstyle\endcsname\relax
  \providecommand{\doi}[1]{https://doi.org/\discretionary{}{}{}#1}\else
  \providecommand{\doi}{https://doi.org/\discretionary{}{}{}\begingroup \urlstyle{rm}\Url}\fi

\bibitem[{Addor et~al.(2017)Addor, Newman, Mizukami, and Clark}]{addor2017camels}
Addor, N., Newman, A.~J., Mizukami, N., and Clark, M.~P.: The CAMELS data set: catchment attributes and meteorology for large-sample studies, Hydrology and Earth System Sciences, 21, 5293--5313, 2017.

\bibitem[{Addor et~al.(2020)Addor, Do, Alvarez-Garreton, Coxon, Fowler, and Mendoza}]{addor2020lsh}
Addor, N., Do, H.~X., Alvarez-Garreton, C., Coxon, G., Fowler, K., and Mendoza, P.~A.: Large-sample hydrology: recent progress, guidelines for new datasets and grand challenges, Hydrological Sciences Journal, 65, 712--725, \doi{10.1080/02626667.2019.1683182}, 2020.

\bibitem[{Allen et~al.(1998)Allen, Pereira, Raes, and Smith}]{pet-guidelines}
Allen, R.~G., Pereira, L.~S., Raes, D., and Smith, M.: Crop evapotranspiration-Guidelines for computing crop water requirements-FAO Irrigation and drainage paper 56, Fao, Rome 300, D05109, \urlprefix\url{https://www.fao.org/4/x0490e/x0490e05.htm}, 1998.

\bibitem[{Alvarez-Garreton et~al.(2018)Alvarez-Garreton, Mendoza, Boisier, Addor, Galleguillos, Zambrano-Bigiarini, Lara, Puelma, Cortes, Garreaud et~al.}]{alvarez2018camels}
Alvarez-Garreton, C., Mendoza, P.~A., Boisier, J.~P., Addor, N., Galleguillos, M., Zambrano-Bigiarini, M., Lara, A., Puelma, C., Cortes, G., Garreaud, R., et~al.: The {CAMELS-CL} dataset: catchment attributes and meteorology for large sample studies--Chile dataset, Hydrology and Earth System Sciences, 22, 5817--5846, 2018.

\bibitem[{Andréassian et~al.(2006)Andréassian, Hall, Chahinian, and Schaake}]{schaake2006mopex}
Andréassian, V., Hall, A., Chahinian, N., and Schaake, J.: Large Sample Basin Experiments for Hydrological Model Parameterization: Results of the Model Parameter Experiment — MOPEX, IAHS Publ., 307, 313--338, \doi{10.1080/13241583.2007.11465316}, 2006.

\bibitem[{Casado~Rodríguez(2023)}]{casado2023ext}
Casado~Rodríguez, J.: CAMELS-ES: Catchment Attributes and Meteorology for Large-Sample Studies – Spain, \doi{10.5281/zenodo.8373020}, 2023.

\bibitem[{Chagas et~al.(2020)Chagas, Chaffe, Addor, Fan, Fleischmann, Paiva, and Siqueira}]{chagas2020camels}
Chagas, V.~B., Chaffe, P.~L., Addor, N., Fan, F.~M., Fleischmann, A.~S., Paiva, R.~C., and Siqueira, V.~A.: {CAMELS-BR}: hydrometeorological time series and landscape attributes for 897 catchments in Brazil, Earth System Science Data, 12, 2075--2096, 2020.

\bibitem[{Clerc-Schwarzenbach et~al.(2024)Clerc-Schwarzenbach, Selleri, Neri, Toth, van Meerveld, and Seibert}]{clerc2024pet}
Clerc-Schwarzenbach, F.~M., Selleri, G., Neri, M., Toth, E., van Meerveld, I., and Seibert, J.: HESS Opinions: A few camels or a whole caravan?, EGUsphere, 2024, 1--29, \doi{10.5194/egusphere-2024-864}, 2024.

\bibitem[{Coxon et~al.(2020)Coxon, Addor, Bloomfield, Freer, Fry, Hannaford, Howden, Lane, Lewis, Robinson et~al.}]{coxon2020camels}
Coxon, G., Addor, N., Bloomfield, J.~P., Freer, J., Fry, M., Hannaford, J., Howden, N.~J., Lane, R., Lewis, M., Robinson, E.~L., et~al.: {CAMELS-GB}: Hydrometeorological time series and landscape attributes for 671 catchments in Great Britain, Earth System Science Data, 12, 2459--2483, 2020.

\bibitem[{Delaigue et~al.(2024)Delaigue, Guimar\~aes, Brigode, G\'enot, Perrin, Soubeyroux, Janet, Addor, and Andr\'eassian}]{camelsfr2024}
Delaigue, O., Guimar\~aes, G.~M., Brigode, P., G\'enot, B., Perrin, C., Soubeyroux, J.-M., Janet, B., Addor, N., and Andr\'eassian, V.: CAMELS-FR dataset: A large-sample hydroclimatic dataset for France to explore hydrological diversity and support model benchmarking, Earth System Science Data Discussions, 2024, 1--27, \doi{10.5194/essd-2024-415}, 2024.

\bibitem[{Fowler et~al.(2021)Fowler, Acharya, Addor, Chou, and Peel}]{fowler2021camels}
Fowler, K.~J., Acharya, S.~C., Addor, N., Chou, C., and Peel, M.~C.: {CAMELS-AUS}: hydrometeorological time series and landscape attributes for 222 catchments in Australia, Earth System Science Data, 13, 3847--3867, 2021.

\bibitem[{Färber et~al.(2023)Färber, Plessow, Kratzert, Addor, Shalev, and Looser}]{farber2023ext}
Färber, C., Plessow, H., Kratzert, F., Addor, N., Shalev, G., and Looser, U.: {GRDC-Caravan: extending the original dataset with data from the Global Runoff Data Centre}, \doi{10.5281/zenodo.10074416}, 2023.

\bibitem[{Helgason(2024)}]{helgason2024ext}
Helgason, H.~B., B.~N.: LamaH-Ice: LArge-SaMple DAta for Hydrology and Environmental Sciences for Iceland, HydroShare, \doi{https://doi.org/10.4211/hs.86117a5f36cc4b7c90a5d54e18161c91}, 2024.

\bibitem[{Helgason and Nijssen(2024)}]{helgason2024lamahice}
Helgason, H.~B. and Nijssen, B.: LamaH-Ice: LArge-SaMple DAta for Hydrology and Environmental Sciences for Iceland, Earth System Science Data, 16, 2741--2771, \doi{10.5194/essd-16-2741-2024}, 2024.

\bibitem[{H\"oge et~al.(2023)H\"oge, Kauzlaric, Siber, Sch\"onenberger, Horton, Schwanbeck, Floriancic, Viviroli, Wilhelm, Sikorska-Senoner, Addor, Brunner, Pool, Zappa, and Fenicia}]{hoge2023camelsch}
H\"oge, M., Kauzlaric, M., Siber, R., Sch\"onenberger, U., Horton, P., Schwanbeck, J., Floriancic, M.~G., Viviroli, D., Wilhelm, S., Sikorska-Senoner, A.~E., Addor, N., Brunner, M., Pool, S., Zappa, M., and Fenicia, F.: CAMELS-CH: hydro-meteorological time series and landscape attributes for 331 catchments in hydrologic Switzerland, Earth System Science Data, 15, 5755--5784, \doi{10.5194/essd-15-5755-2023}, 2023.

\bibitem[{Höge et~al.(2023)Höge, Kauzlaric, Siber, Schönenberger, Horton, Schwanbeck, Floriancic, Viviroli, Wilhelm, Sikorska-Senoner, Addor, Brunner, Pool, Zappa, and Fenicia}]{hoge2023ext}
Höge, M., Kauzlaric, M., Siber, R., Schönenberger, U., Horton, P., Schwanbeck, J., Floriancic, M.~G., Viviroli, D., Wilhelm, S., Sikorska-Senoner, A.~E., Addor, N., Brunner, M., Pool, S., Zappa, M., and Fenicia, F.: {Catchment attributes and hydro-meteorological time series for large-sample studies across hydrologic Switzerland (CAMELS-CH)}, \doi{10.5281/zenodo.7928595}, 2023.

\bibitem[{Klingler et~al.(2021)Klingler, Schulz, and Herrnegger}]{klingler2021lamah}
Klingler, C., Schulz, K., and Herrnegger, M.: {LamaH-CE}: LArge-SaMple DAta for Hydrology and Environmental Sciences for Central Europe, Earth System Science Data, 13, 4529--4565, 2021.

\bibitem[{Koch(2022)}]{koch_2022_ext}
Koch, J.: {Caravan extension Denmark - Danish dataset for large-sample hydrology}, \doi{10.5281/zenodo.7396466}, 2022.

\bibitem[{Kratzert et~al.(2021)Kratzert, Klotz, Hochreiter, and Nearing}]{kratzert2023synergy}
Kratzert, F., Klotz, D., Hochreiter, S., and Nearing, G.~S.: A note on leveraging synergy in multiple meteorological data sets with deep learning for rainfall--runoff modeling, Hydrology and Earth System Sciences, 25, 2685--2703, \doi{10.5194/hess-25-2685-2021}, 2021.

\bibitem[{Kratzert et~al.(2023)Kratzert, Nearing, Addor, Erickson, Gauch, Gilon, Gudmundsson, Hassidim, Klotz, Nevo et~al.}]{kratzert2023caravan}
Kratzert, F., Nearing, G., Addor, N., Erickson, T., Gauch, M., Gilon, O., Gudmundsson, L., Hassidim, A., Klotz, D., Nevo, S., et~al.: Caravan-A global community dataset for large-sample hydrology, Scientific Data, 10, 61, 2023.

\bibitem[{Lam et~al.(2023)Lam, Sanchez-Gonzalez, Willson, Wirnsberger, Fortunato, Alet, Ravuri, Ewalds, Eaton-Rosen, Hu, Merose, Hoyer, Holland, Vinyals, Stott, Pritzel, Mohamed, and Battaglia}]{graphcast2023}
Lam, R., Sanchez-Gonzalez, A., Willson, M., Wirnsberger, P., Fortunato, M., Alet, F., Ravuri, S., Ewalds, T., Eaton-Rosen, Z., Hu, W., Merose, A., Hoyer, S., Holland, G., Vinyals, O., Stott, J., Pritzel, A., Mohamed, S., and Battaglia, P.: Learning skillful medium-range global weather forecasting, Science, 382, 1416--1421, \doi{10.1126/science.adi2336}, 2023.

\bibitem[{Liu et~al.(2024)Liu, Koch, Stisen, Troldborg, H{\o}jberg, Thodsen, Hansen, and Schneider}]{liu2024camelsdk}
Liu, J., Koch, J., Stisen, S., Troldborg, L., H{\o}jberg, A.~L., Thodsen, H., Hansen, M. F.~T., and Schneider, R. J.~M.: CAMELS-DK: Hydrometeorological Time Series and Landscape Attributes for 3330 Catchments in Denmark, Earth System Science Data Discussions, 2024, 1--30, \doi{10.5194/essd-2024-292}, 2024.

\bibitem[{Loritz et~al.(2024)Loritz, Dolich, Acu\~na Espinoza, Ebeling, Guse, G\"otte, Hassler, Hauffe, Heidb\"uchel, Kiesel, M\"alicke, M\"uller-Thomy, St\"olzle, and Tarasova}]{loritz2023camelsde}
Loritz, R., Dolich, A., Acu\~na Espinoza, E., Ebeling, P., Guse, B., G\"otte, J., Hassler, S.~K., Hauffe, C., Heidb\"uchel, I., Kiesel, J., M\"alicke, M., M\"uller-Thomy, H., St\"olzle, M., and Tarasova, L.: CAMELS-DE: hydro-meteorological time series and attributes for 1555 catchments in Germany, Earth System Science Data Discussions, 2024, 1--30, \doi{10.5194/essd-2024-318}, 2024.

\bibitem[{Mangukiya et~al.(2024)Mangukiya, Kumar, Dey, Sharma, Bejagam, Mujumdar, and Sharma}]{mangukiya2024camelsin}
Mangukiya, N.~K., Kumar, K.~B., Dey, P., Sharma, S., Bejagam, V., Mujumdar, P.~P., and Sharma, A.: CAMELS-INDIA: hydrometeorological time series and catchment attributes for 472 catchments in Peninsular India, Earth System Science Data Discussions, 2024, 1--43, \doi{10.5194/essd-2024-379}, 2024.

\bibitem[{Morin(2023)}]{morin2023ext}
Morin, E.: {Caravan extension Israel - Israel dataset for large-sample hydrology}, \doi{10.5281/zenodo.7758516}, 2023.

\bibitem[{Mu{\~n}oz-Sabater et~al.(2021)Mu{\~n}oz-Sabater, Dutra, Agust{\'\i}-Panareda, Albergel, Arduini, Balsamo, Boussetta, Choulga, Harrigan, Hersbach et~al.}]{munoz2021era5}
Mu{\~n}oz-Sabater, J., Dutra, E., Agust{\'\i}-Panareda, A., Albergel, C., Arduini, G., Balsamo, G., Boussetta, S., Choulga, M., Harrigan, S., Hersbach, H., et~al.: ERA5-Land: A state-of-the-art global reanalysis dataset for land applications, Earth System Science Data, 13, 4349--4383, 2021.

\bibitem[{Nearing et~al.(2024)Nearing, Cohen, Dube, Gauch, Gilon, Harrigan, Hassidim, Klotz, Kratzert, Metzger, Nevo, Pappenberger, Prudhomme, Shalev, Shenzis, Tekalign, Weitzner, and Matias}]{nearing2024nature}
Nearing, G., Cohen, D., Dube, V., Gauch, M., Gilon, O., Harrigan, S., Hassidim, A., Klotz, D., Kratzert, F., Metzger, A., Nevo, S., Pappenberger, F., Prudhomme, C., Shalev, G., Shenzis, S., Tekalign, T.~Y., Weitzner, D., and Matias, Y.: Global prediction of extreme floods in ungauged watersheds, Nature, \doi{10.1038/s41586-024-07145-1}, 2024.

\bibitem[{Newman et~al.(2015)Newman, Clark, Sampson, Wood, Hay, Bock, Viger, Blodgett, Brekke, Arnold, Hopson, and Duan}]{newman2015camels}
Newman, A.~J., Clark, M.~P., Sampson, K., Wood, A., Hay, L.~E., Bock, A., Viger, R.~J., Blodgett, D., Brekke, L., Arnold, J.~R., Hopson, T., and Duan, Q.: Development of a large-sample watershed-scale hydrometeorological data set for the contiguous USA: data set characteristics and assessment of regional variability in hydrologic model performance, Hydrology and Earth System Sciences, 19, 209--223, \doi{10.5194/hess-19-209-2015}, 2015.

\bibitem[{Singer et~al.()Singer, Asfaw, Rosolem, Cuthbert, Miralles, MacLeod, Quichimbo, and Michaelides}]{singer2021fao}
Singer, M.~B., Asfaw, D.~T., Rosolem, R., Cuthbert, M.~O., Miralles, D.~G., MacLeod, D., Quichimbo, E.~A., and Michaelides, K.: Hourly potential evapotranspiration at 0.1° resolution for the global land surface from 1981-present, Scientific Data, \doi{10.1038/s41597-021-01003-9}.

\bibitem[{Xie et~al.(2007)Xie, Yatagai, Chen, Hayasaka, Fukushima, Liu, and Yang}]{cpc2007}
Xie, P., Yatagai, A., Chen, M., Hayasaka, T., Fukushima, Y., Liu, C., and Yang, S.: A gauge-based analysis of daily precipitation over East Asia, Hydrometeorol, \doi{10.1038/s41586-024-07145-1}, 2007.

\end{thebibliography}

\end{document}